\documentclass[letterpaper, 10 pt, journal, twoside]{IEEEtran}  
\IEEEoverridecommandlockouts                              
\usepackage{cite}
\usepackage{booktabs}  
\usepackage{amsmath}
\usepackage{graphicx}
\usepackage{float}
\usepackage{balance}

\title{\LARGE \bf
A Two-Layer Electrostatic Film Actuator with High Actuation Stress and Integrated Brake}

\author{Huacen Wang, and Hongqiang Wang,~\IEEEmembership{Member,~IEEE,}
\thanks{This work was supported in part by the Natural Science Foundation of China under Grant 52275021 and 52450323, the Natural Science Foundation of Guangdong Province of China under Grant 2024A1515010183, and in part by Guangdong Provincial Research Support Program 2019QN01Z733. (Corresponding author: Hongqiang Wang.)}
\thanks{Huacen Wang is with the Department of Mechanical and Energy Engineering, Southern University of Science and Technology, Shenzhen 518055, China.}
\thanks{Hongqiang Wang is with the Department of Mechanical and Energy Engineering, Southern University of Science and Technology, Shenzhen 518055, China(e-mail: wanghq6@sustech.edu.cn).}}

\begin{document}

\maketitle
\thispagestyle{empty}
\pagestyle{empty}

\begin{abstract}

Robotic systems driven by conventional motors often suffer from challenges such as large mass, complex control algorithms, and the need for additional braking mechanisms, which limit their applications in lightweight and compact robotic platforms. Electrostatic film actuators offer several advantages, including thinness, flexibility, lightweight construction, and high open-loop positioning accuracy. However, the actuation stress exhibited by conventional actuators in air still needs improvement, particularly for the widely used three-phase electrode design. To enhance the output performance of actuators, this paper presents a two-layer electrostatic film actuator with an integrated brake. By alternately distributing electrodes on both the top and bottom layers, a smaller effective electrode pitch is achieved under the same fabrication constraints, resulting in an actuation stress of approximately 241~N/m$^2$, representing a 90.5\% improvement over previous three-phase actuators operating in air. Furthermore, its integrated electrostatic adhesion mechanism enables load retention under braking mode. Several demonstrations, including a tug-of-war between a conventional single-layer actuator and the proposed two-layer actuator, a payload operation, a one-degree-of-freedom robotic arm, and a dual-mode gripper, were conducted to validate the actuator's advantageous capabilities in both actuation and braking modes.

\end{abstract}

\section{INTRODUCTION}

With the development of industrial automation and intelligent manufacturing technologies, the demand for robotic systems in human society has been steadily increasing. However, most robotic systems are based on magnetic motors combined with rigid transmission mechanisms, resulting in substantial mass ~\cite{axinte2011free, liu2020novel, mei2023mobile} and limiting their applications in human-machine interaction and miniature robotics. Furthermore, to withstand static loads while maintaining a desired position, magnetic motors typically require external braking components ~\cite{wang2021optimized, dai2021design, wang2023optimal} or closed-loop control systems ~\cite{de2005pd, calanca2015review, abu2020variable}, increasing the system complexity. Soft actuators, such as dielectric elastomers \cite{cao2018untethered, tang2022pipeline}, shape memory alloys\cite{yang202088, an2023active}, and pneumatic actuators \cite{brown2010universal, ilievski2011soft, yang2025soft}, have been widely applied in miniature robots, wearable devices, and grippers in recent years. However, their viscoelastic materials mostly result in limited accuracy and a long response time \cite{rich2018untethered,el2020soft}.

Electrostatic film actuators (EFAs), driven by electrostatic forces generated between microscale comb electrodes embedded within the polymer film insulation\cite{niino1997dual,choi2024linear}, offer advantages including thinness, flexibility, lightweight construction, and high open-loop positioning accuracy. In addition, they can provide sufficient actuation velocity and force \cite{zhang2022high}, making them suitable for applications in human-machine interaction \cite{hosobata2014transparent,hosobata20132} and bio-inspired robotics \cite{wang2013thin,wang2014crawler,wang2017analyses}. Due to their operation under high driving voltage amplitude but relatively low current, EFAs also exhibit resistance to strong magnetic interference\cite{gassert2006actuation,rajendra2008motion}.

However, in conventional flexible printed circuit manufacturing processes, both the electrode width and spacing are constrained by the same minimum feature size. Conventional EFAs arrange all electrodes within a single electrode layer \cite{wang2021modeling,qu2021analyses,zhang2023position}, limiting further reduction of effective electrode pitch (the sum of the width and spacing) and thus restricting the achievable actuation force \cite{wang2021biologically}. Moreover, the asymmetry in the distance between the electrode layer and the surfaces results in unbalanced actuation forces on both sides of the film, which degrades performance when multiple layers are stacked \cite{choi2024linear,niino1994electrostatic}. A four-phase two-layer EFA design has previously been reported to achieve higher actuation stress and applied to mesoscale artificial muscles \cite{wang2021biologically}. However, it requires more precise fabrication techniques and more driving equipment. Accordingly, the development of electrode design capable of enhancing the actuation force density within the constraints of existing fabrication processes remains a challenge in advancing EFA performance.

\begin{figure}[htbp]
  \centering
  \includegraphics[width=8cm]{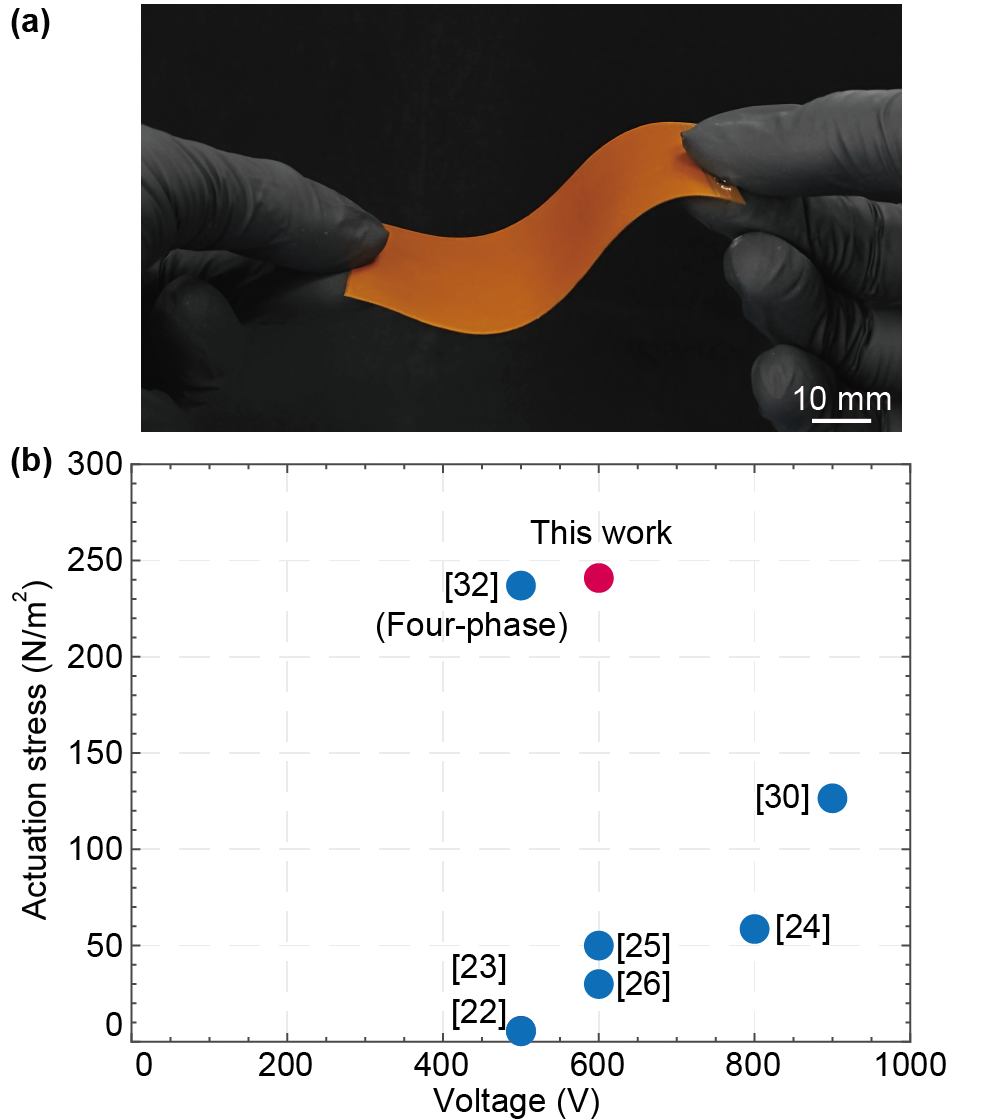}
  \setlength{\abovecaptionskip}{-0.1cm}
  \caption{The two-layer electrostatic film actuator. (a) Demonstration of the flexible and thin structure of the actuator. (b) Comparison of the actuation stress of the proposed two-layer three-phase actuator with previously reported actuators operating in air.}
  \label{fig1}
\end{figure}

The contributions of this work are as follows. In this work, we propose a two-layer EFA (Fig.~\ref{fig1}a), where electrodes are alternately arranged more densely across the upper and lower layers. This design enables a smaller effective electrode pitch under the same fabrication conditions, resulting in higher actuation stress compared to conventional three-phase EFAs under the same driving voltage amplitude, as shown in Fig.~\ref{fig1}b. In addition, the larger spacing between electrodes in the same layer improves the breakdown strength compared to the four-phase EFA, achieving actuation stresses up to approximately 241 N/m² and allowing the actuator to operate with a payload of approximately 68.2 times its own weight. Benefiting from the symmetric two-layer electrode design, the actuation film generates equal output forces on both sides, thereby facilitating force enhancement through stacking multiple films. Furthermore, the electrode coverage ratio reaches as high as 83.3\% (with half of the electrodes located on the other layer), enabling the actuator to generate up to 12.0 N of electrostatic adhesion force for braking functionality. Based on this two-layer EFA with integrated brake, we demonstrate its application in driving a one-degree-of-freedom (DOF) robotic arm capable of handling heavier loads during active braking. Moreover, a dual-mode gripper is developed, utilizing the EFA's actuation mode for object grasping and its braking mode for object holding.

\begin{figure}[htbp]
  \centering
  \includegraphics[width=8cm]{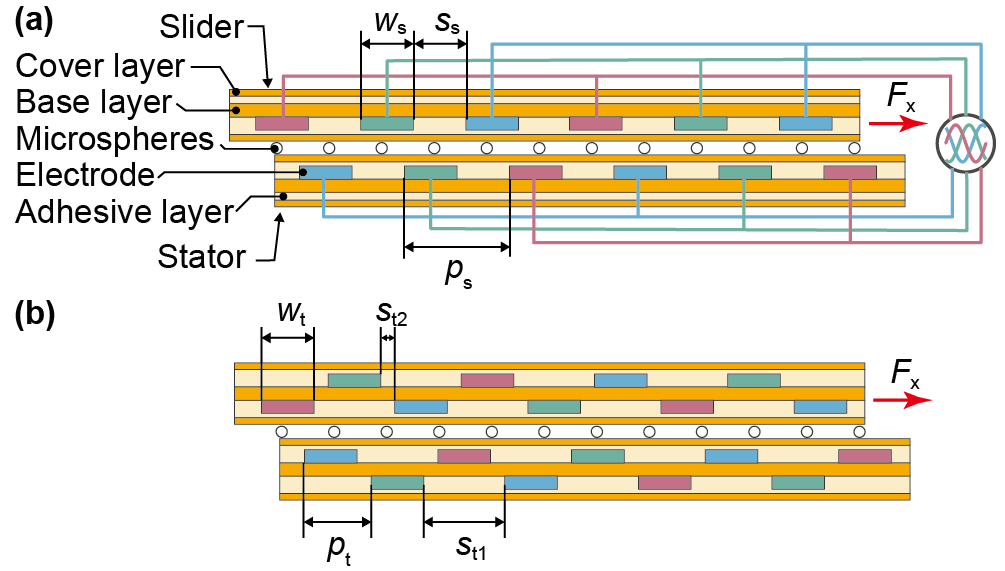}
  \setlength{\abovecaptionskip}{-0.1cm}
  \caption{The principle and design of (a) conventional single-layer EFA and (b) the two-layer EFA.}
  \label{fig2}
\end{figure}

The structure of this paper is organized as follows. Section II presents the working principle and design of the two-layer EFA. Section III describes the fabrication process and the microstructure of the actuator electrode. Section IV details the experimental setups and evaluates the actuator's performance. Section V demonstrates the applications of a robotic arm and a gripper. Section VI concludes the paper and discusses future work.

\section{ACTUATOR DESIGN AND ANALYSES}

\subsection{Principle and design of the two-layer EFA}

The design of the slider and stator in conventional single-layer EFAs is shown in Fig.~\ref{fig2}a. Three-phase electrodes are arranged in parallel on the base layer and electrically insulated by the cover layers through adhesive layers. In terms of electrode parameters, $w_s$ denotes the electrode width, $s_s$ represents the electrode spacing, and $p_s$, defined as the sum of the width and spacing ($p_s = w_s + s_s$), is referred to as the effective electrode pitch. Due to the identical minimum size limitation of the fabrication process for both width and spacing, $w_s$ is typically designed to be equal to $s_s$. As a normal adhesion force is generated between the slider and the stator during operation, glass beads (with a diameter of 20 to 30~$\mu$m) are uniformly applied to the contact surfaces of the films to reduce friction. When three-phase sinusoidal voltages with initial phase differences of $120^\circ$ are applied to the electrodes, a horizontal driving force $F_x$ is generated on the slider, which can be expressed as~\cite{zhang2022high}:
\begin{equation}
F_x = K_a v_a^2 \sin(\theta_e - 2\phi),
\end{equation}
where $K_a = k_a A_e$ is the actuation force constant with units of N$\cdot$V$^{-2}$ of the actuator. $k_a$ is a material-dependent force coefficient with units of N$\cdot$m$^{-2}$$\cdot$V$^{-2}$. $A_e$ is the effective actuation area of the actuator. $v_a$ is the amplitude of the driving voltages. $\theta_e = 2\pi x_\text{sl}/3p$ represents the electric angle of the slider position $x_\text{sl}$ along the X direction, with $p$ denoting the effective pitch. The parameter $\phi$ is the excitation phase of the applied voltage.

\begin{figure*}[htbp]
  \centering
  \includegraphics[width=16cm]{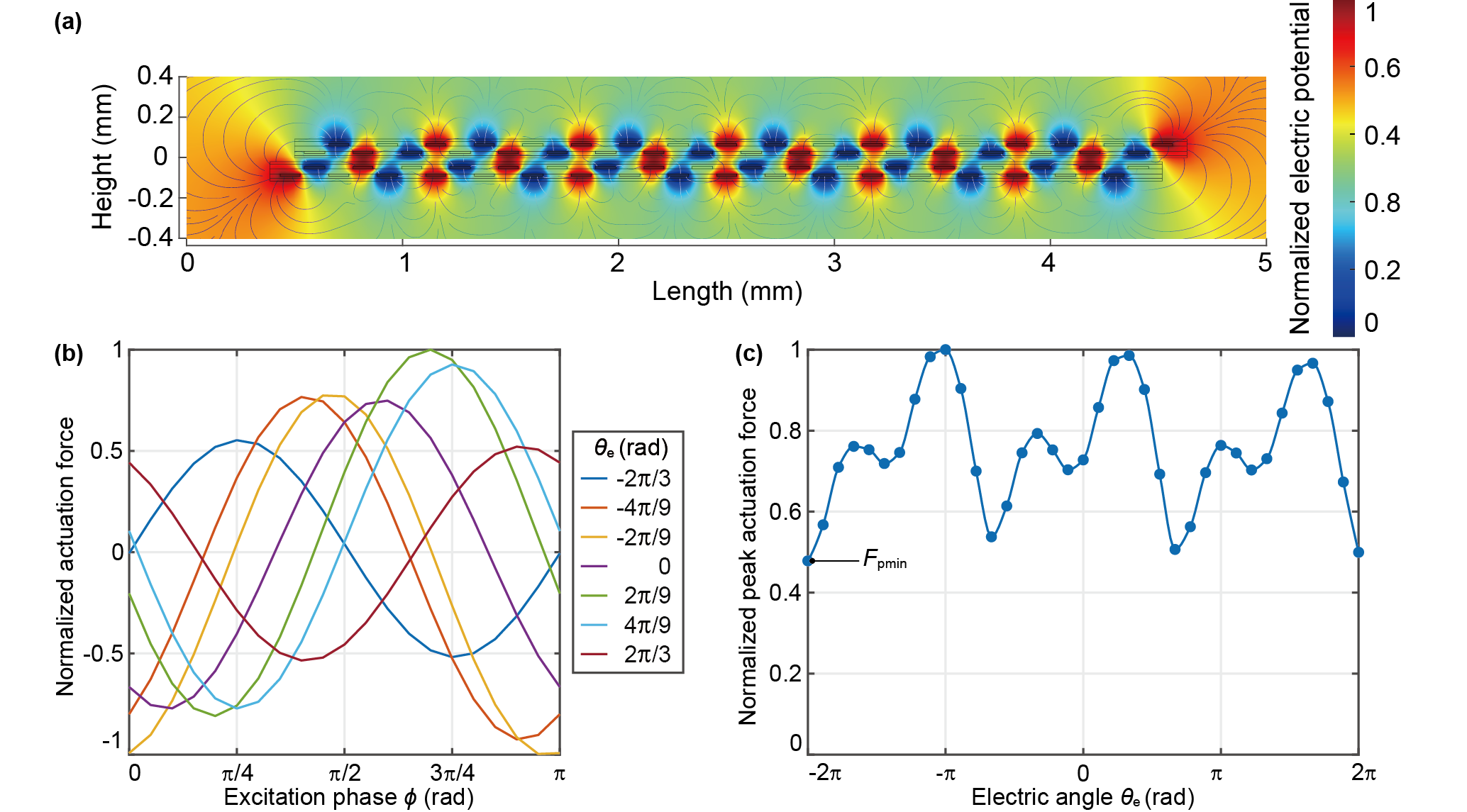}
  \setlength{\abovecaptionskip}{-0.1cm}
  \caption{Analysis of the two-layer EFA. (a) The normalized electric potential distribution of Actuator 1. (b) Relationship between normalized actuation force and excitation phase in different electric angles. (c) Relationship between normalized peak actuation force and electric angle.}
  \label{fig3}
  \vspace{-0.5cm}
\end{figure*}

\begin{figure}[htbp]
  \centering
  \includegraphics[width=8cm]{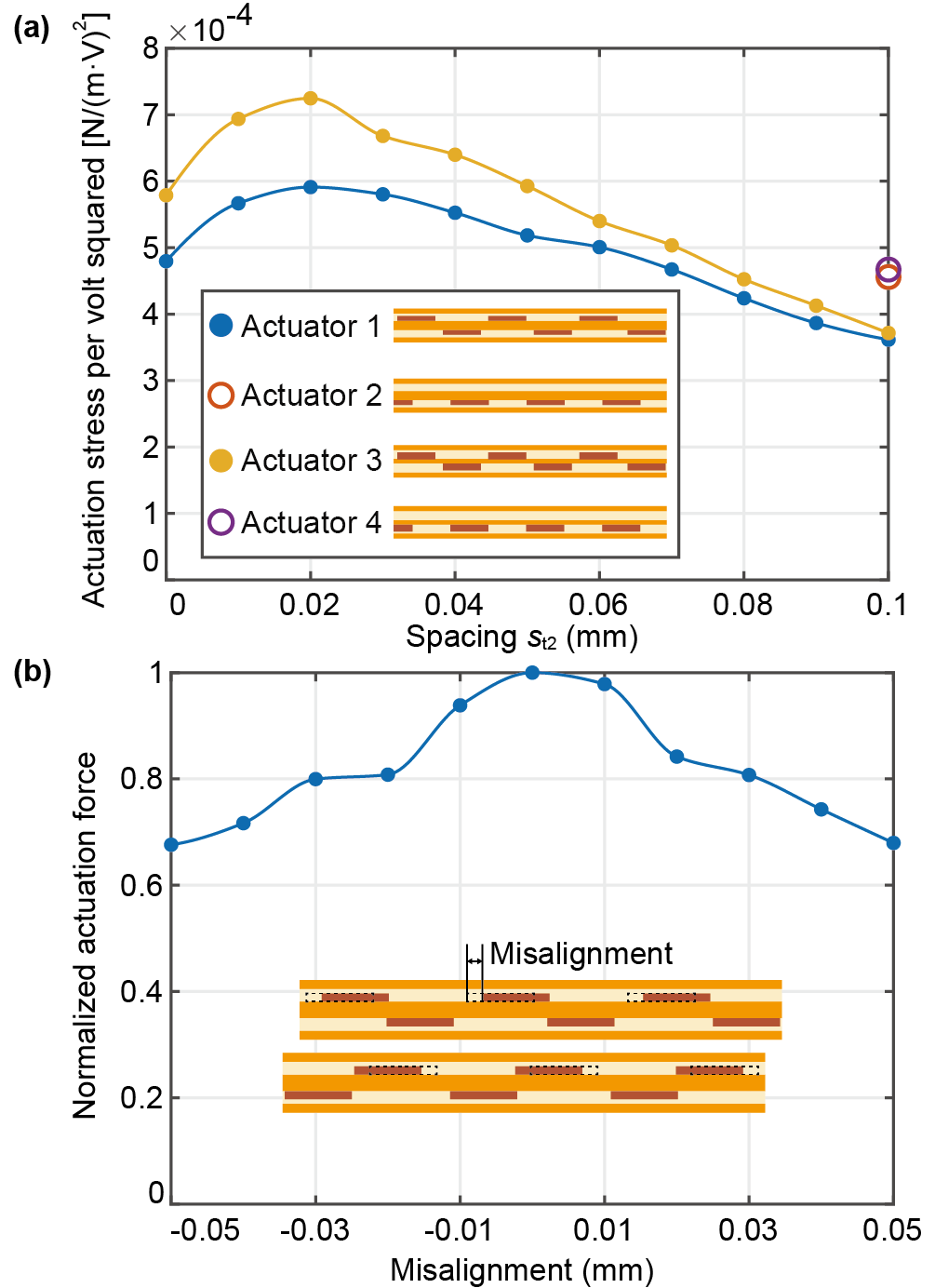}
  \setlength{\abovecaptionskip}{-0.1cm}
  \caption{ The influence of design and fabrication parameters on the actuation force. (a) Relationship between actuation stress per volt squared and spacing. (b) Relation between normalized actuation force and misalignment.}
  \label{fig4}
  \vspace{-0.5cm}
\end{figure}

\begin{figure*}[htbp]
  \centering
  \includegraphics[width=16cm]{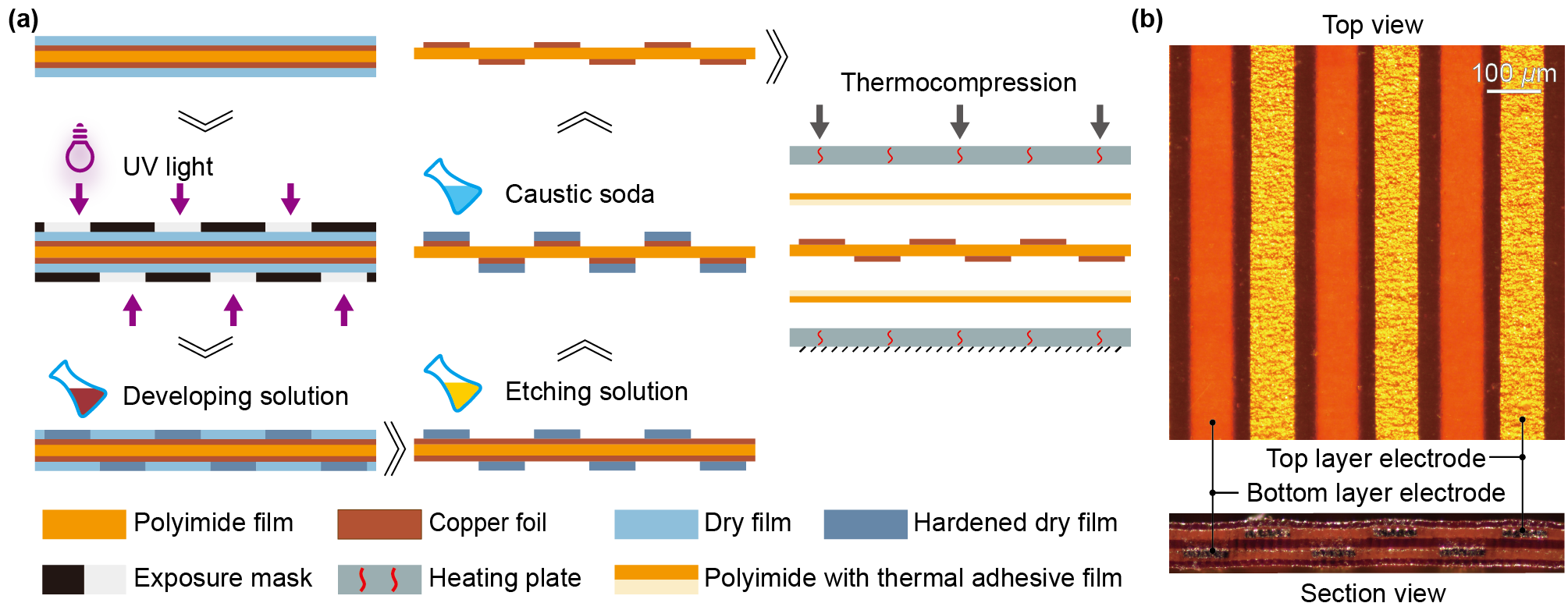}
  \setlength{\abovecaptionskip}{-0.1cm}
  \caption{Fabrication of the two-layer actuator. (a) The illustration of the fabrication process. (b) The top and sectional views of the electrodes.}
  \label{fig5}
  \vspace{-0.5cm}
\end{figure*}

Since EFAs operate in synchronous motion, their actuation velocity can be expressed as~\cite{yamamoto2006modeling}:
\begin{equation}
v = 6pf,
\end{equation}
where $f$ is the driving frequency of the applied voltage.

The design of the proposed two-layer EFA is illustrated in Fig.~\ref{fig2}b. The electrodes are alternately arranged on both the top and bottom layers. In this configuration, $s_{t1}$ denotes the electrode spacing on the same layer, $s_{t2}$ represents the adjacent electrode spacing on different layers, and the effective electrode pitch $p_t = w_t + s_{t2}$. When utilizing an identical fabrication process for both single-layer and two-layer designs, the electrode widths remain the same ($w_t = w_s$). Since $s_{t2}$ can be adjusted without requiring higher fabrication precision, a smaller effective pitch can be achieved when $s_{t2}$ is between 0 and $s_s$, resulting in $p_t$ smaller than $p_s$. Meanwhile, the same-layer electrode spacing satisfies $s_{t1} = w_t + 2s_{t2}$, which is larger than $s_s$ ($s_s = w_t = w_s$), resulting in a higher breakdown strength and thus enabling a higher maximum actuation force of the EFA.

In addition to actuation, the electrostatic adhesion capability of the two-layer EFA can be utilized for braking. During adhesion operation, all slider electrodes are connected to a square-wave AC voltage, while all stator electrodes are grounded. Under this configuration, the system can be approximated by a parallel-plate capacitor model, and the normal adhesion force can be calculated as ~\cite{xiong2022so}:
\begin{equation}
F_n = K_b v_b^2,
\end{equation}
where $K_b$ is the adhesion force constant with units of N$\cdot$V$^{-2}$ based on the materials and dimensions of the electrostatic adhesive pad, $v_b$ is the amplitude of the driving voltage for adhesion. Since the actuation force operates tangentially, the corresponding tangential adhesion force used for braking can be expressed as:
\begin{equation}
F_t = \mu_b F_n = \mu_b K_b v_b^2,
\end{equation}
where $\mu_b$ is the friction coefficient between the slider and the stator.

\subsection{Finite element analysis}
Finite element analysis was employed to investigate the effects of electrode configurations and fabrication parameters on the actuation forces of EFAs. In the simulation, the slider and stator were positioned inside an air-filled enclosure, with an air gap of 20~$\mu$m and a relative permittivity of 1. The base layer and cover layer materials were set as polyimide with a relative permittivity of 3.4, while the adhesive layer was set as acrylic adhesive with a relative permittivity of 3.5. The electrode material was set as copper. The thicknesses of the cover layer and adhesive layer were set to 13~$\mu$m and 2~$\mu$m, respectively. As summarized in Table~\ref{tab:parameters}, simulations were carried out for four types of actuators to compare their actuation forces under different design parameters. Specifically, combinations of electrode layer thickness and base layer thickness were selected based on two kinds of commercially available double-sided copper-clad laminates.

\begin{table}[htbp]
\centering\setlength{
\abovecaptionskip}{-0.1cm}
\caption{The parameters of four types of actuators in simulations.}
\label{tab:parameters}
\begin{tabular}{llcc}
\toprule
\textbf{Actuator} & \textbf{Single/two-layer} & \multicolumn{2}{c}{\textbf{Thickness ($\mu$m)}} \\
\cmidrule(lr){3-4}
& & \textbf{Electrode layer} & \textbf{Base layer} \\
\midrule
Actuator 1 & Two    & 12   & 25   \\
Actuator 2 & Single & 12   & 25   \\
Actuator 3 & Two    & 18   & 12.5 \\
Actuator 4 & Single & 18   & 12.5 \\
\bottomrule
\end{tabular}
\end{table}

By applying driving voltages to the electrodes of each phase in the simulation, as shown in Fig.~\ref{fig3}a, the electric potential distribution of the two-layer EFA (corresponding to Actuator~1 in Table~1, with $w_t = 100~\mu\text{m}$ and $s_{t2} = 20~\mu\text{m}$) was obtained. This further yielded the relationship between the normalized actuation force and the excitation phase at different electric angles within two effective pitches (the period of the two-layer actuator spans six effective pitches), as shown in Fig.~\ref{fig3}b. The relationship between the peak actuation force and electric angle is presented in Fig.~\ref{fig3}c, where $F_\text{pmin}$ denotes the maximum achievable normalized actuation force during continuous motion.

To identify optimal design parameters for the two-layer EFA, the influence of the spacing $s_{t2}$ on actuation force was further investigated. In this comparison, all electrode widths were fixed at $100~\mu\text{m}$. For the single-layer EFAs (Actuators 2 and 4), the electrode spacing $s_s$ was set to $100~\mu\text{m}$. The relationship between actuation stress per volt squared and spacing $s_{t2}$ is shown in Fig.~\ref{fig4}a. The two-layer EFAs exhibited maximum actuation forces when $s_{t2}$ was set to $20~\mu\text{m}$, and were much higher than those of the single-layer counterparts (the force of Actuator~1 is 29.7\% larger than that of Actuator~2).

In the fabrication process of the two-layer EFA, electrode misalignment between electrodes in the top and bottom layers may occur due to processing error. The influence of the misalignment on the actuation force was analyzed through simulation. The relationship between normalized actuation force and electrode misalignment is shown in Fig.~\ref{fig4}b (corresponding to Actuator~1 with $w_t = 100~\mu\text{m}$ and $s_{t2} = 20~\mu\text{m}$). As the value of misalignment increases, the actuation force decreases significantly. Therefore, the misalignment should be controlled within $10~\mu\text{m}$ during fabrication to ensure optimal actuator performance.

\section{FABRICATION}
The fabrication procedure of the two-layer EFA film is shown in Fig.~\ref{fig5}a. Firstly, a double-sided copper-clad laminate (comprising a $25~\mu\text{m}$ polyimide base layer and two $12~\mu\text{m}$ copper layers) served as the substrate for the actuator. Then, negative-type dry films were laminated to both sides of the substrate. Ultraviolet (UV) light was subsequently projected through exposure masks containing the designed electrode layouts. The exposed portions of the photosensitive dry films were hardened, forming insoluble resist layers, while the unexposed regions were removed by a developing solution (1\% Na$_2$CO$_3$). The uncovered copper was then etched away using an etching solution (CuCl$_2$ and HCl), leaving the designed electrode patterns. After etching, the residual hardened dry films were stripped with a caustic soda solution (3\% NaOH), exposing the patterned flexible circuits. For insulation, two coverlays (consisting of a $13~\mu\text{m}$ polyimide layer and a $13~\mu\text{m}$ thermal adhesive layer) were laminated onto both sides of the etched substrate using a thermocompression at $185^\circ$C and 2.3~MPa for 3 hours.

The final fabricated two-layer EFA film measured $100~\text{mm} \times 50~\text{mm}$ in length and width, with a total thickness of $90~\mu\text{m}$ and a mass of 0.88~g. The effective electrode area was $92.8~\text{mm} \times 45.6~\text{mm}$. The microscopic top view and the section view of the fabricated electrodes are shown in Fig.~\ref{fig5}b. Inside the actuation film, the electrodes were alternately distributed on the top and bottom layers. The fabricated copper electrodes had an average width of approximately $85~\mu\text{m}$ and a spacing of approximately $35~\mu\text{m}$. These dimensions slightly deviate from the design values of $100~\mu\text{m}$ width and $20~\mu\text{m}$ spacing, which may be attributed to over-etching during the fabrication process.

\begin{figure}[htbp]
  \centering
  \includegraphics[width=8cm]{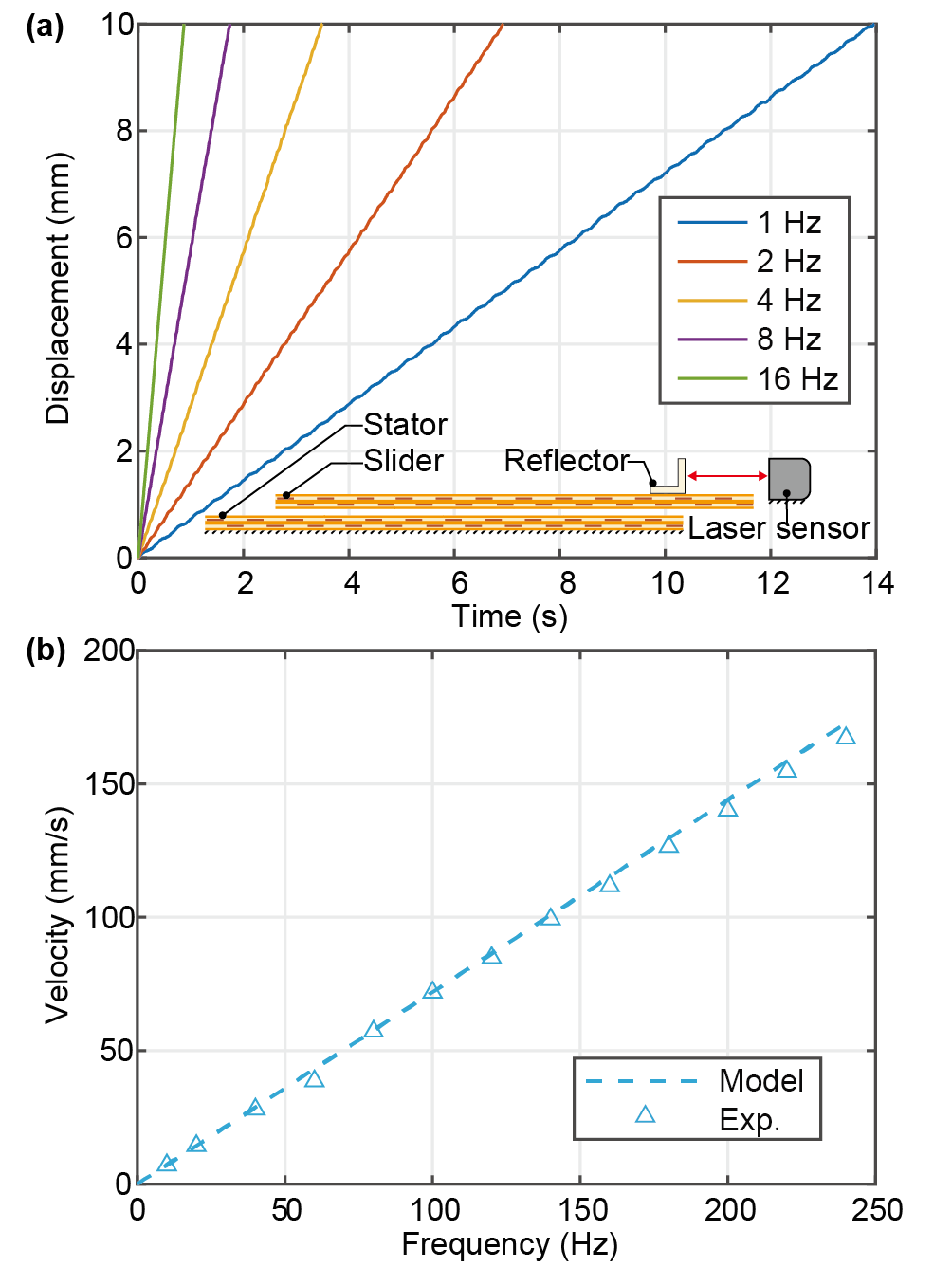}
  \setlength{\abovecaptionskip}{-0.1cm}
  \caption{The velocity of the actuator. (a) Schematic of the experimental setup and results of displacements in different driving frequencies. (b) Results of velocity in driving frequencies.}
  \label{fig6}
  \vspace{-0.5cm}
\end{figure}
\begin{figure}[htbp]
  \centering
  \includegraphics[width=8cm]{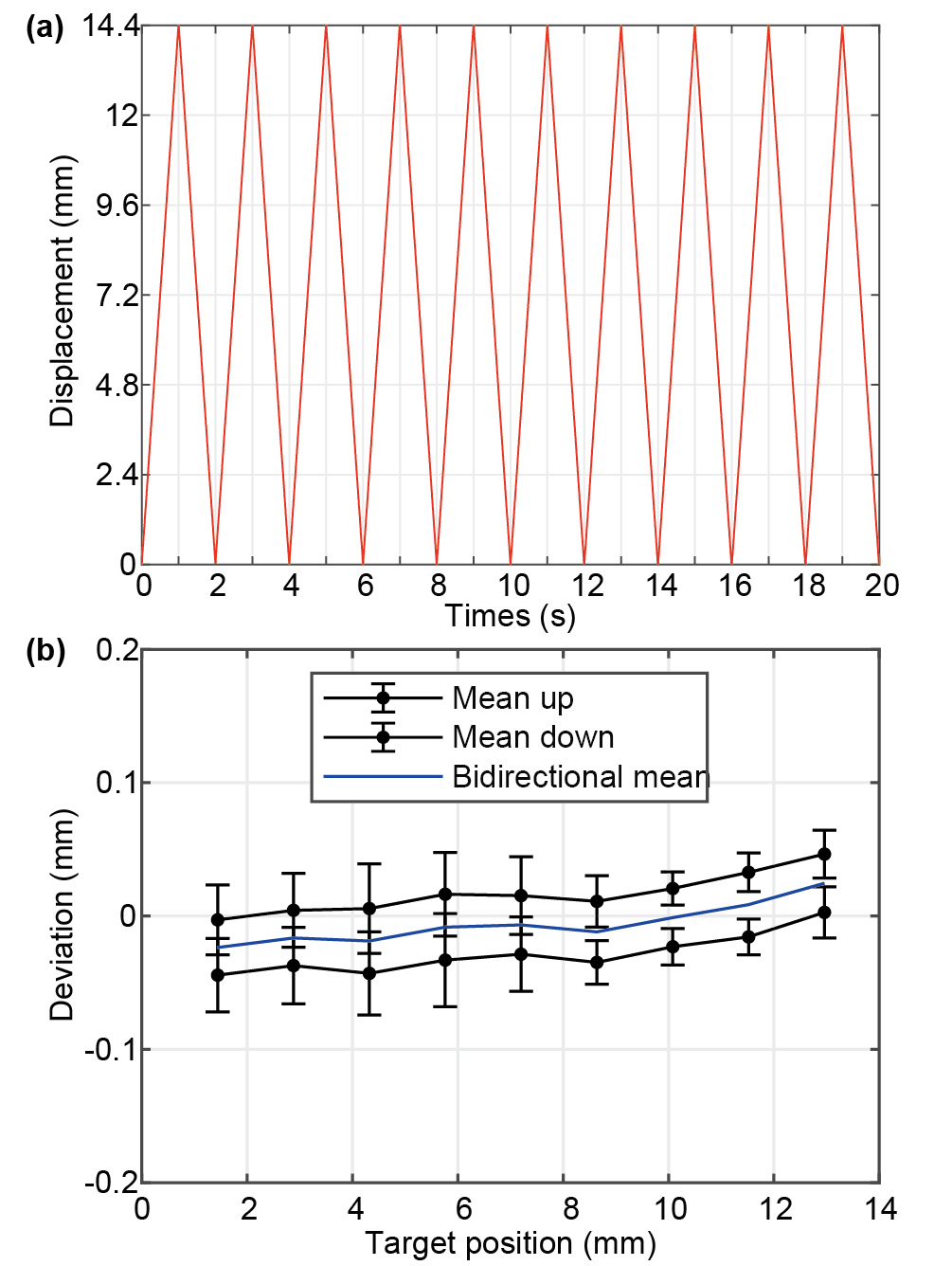}
  \setlength{\abovecaptionskip}{-0.1cm}
  \caption{The positioning accuracy of the actuator. (a) Results of displacement data for 10 cycles of reciprocal movements. (b) Results of the positioning error.}
  \label{fig7}
  \vspace{-0.5cm}
\end{figure}
\section{EXPERIMENT RESULTS}
We experimentally explored the velocity, positioning accuracy, actuation force, and braking force of the two-layer EFA. In these experiments, control signals were generated by a controller board (SCB-68A, National Instruments) and amplified 1000 times using three high-voltage amplifiers (Model 615-3, Trek) to drive the actuator. Sensor data were collected using analog signal acquisition boards (TB-4309 and TB-4330, National Instruments).

\subsection{Results of velocity and positioning accuracy} 
In the displacement measurement setup shown in Fig.~\ref{fig6}a, the stator was fixed on an acrylic plate, while a reflector was mounted on the slider for displacement sensing via a laser displacement sensor (HG-C1200, Panasonic). In the displacement tests, the voltage amplitude was set to 500~V$_\text{0-p}$. The displacements of the actuator under different driving frequencies are presented in Fig.~\ref{fig6}a. As shown in Fig.~\ref{fig6}b, the actuator velocity exhibited a linear relationship with the driving frequency, which is consistent with the theoretical model. The maximum velocity reached approximately 167~mm/s at a 240~Hz driving frequency. To evaluate linear positioning accuracy, the slider was driven to reciprocate over a 14.4~mm stroke at a driving frequency of 20~Hz for ten cycles, as shown in Fig.~\ref{fig7}a. According to the results and based on the accuracy evaluation method of the Physik Instrumente (PI)~\cite{Grabowski2020}, the mean bidirectional positioning accuracy of the actuator was 0.048~mm, with a bidirectional repeatability accuracy of 0.035~mm as shown in Fig.~\ref{fig7}b.

\subsection{Results of the actuation force}
\begin{figure}[htbp]
  \centering
  \includegraphics[width=8cm]{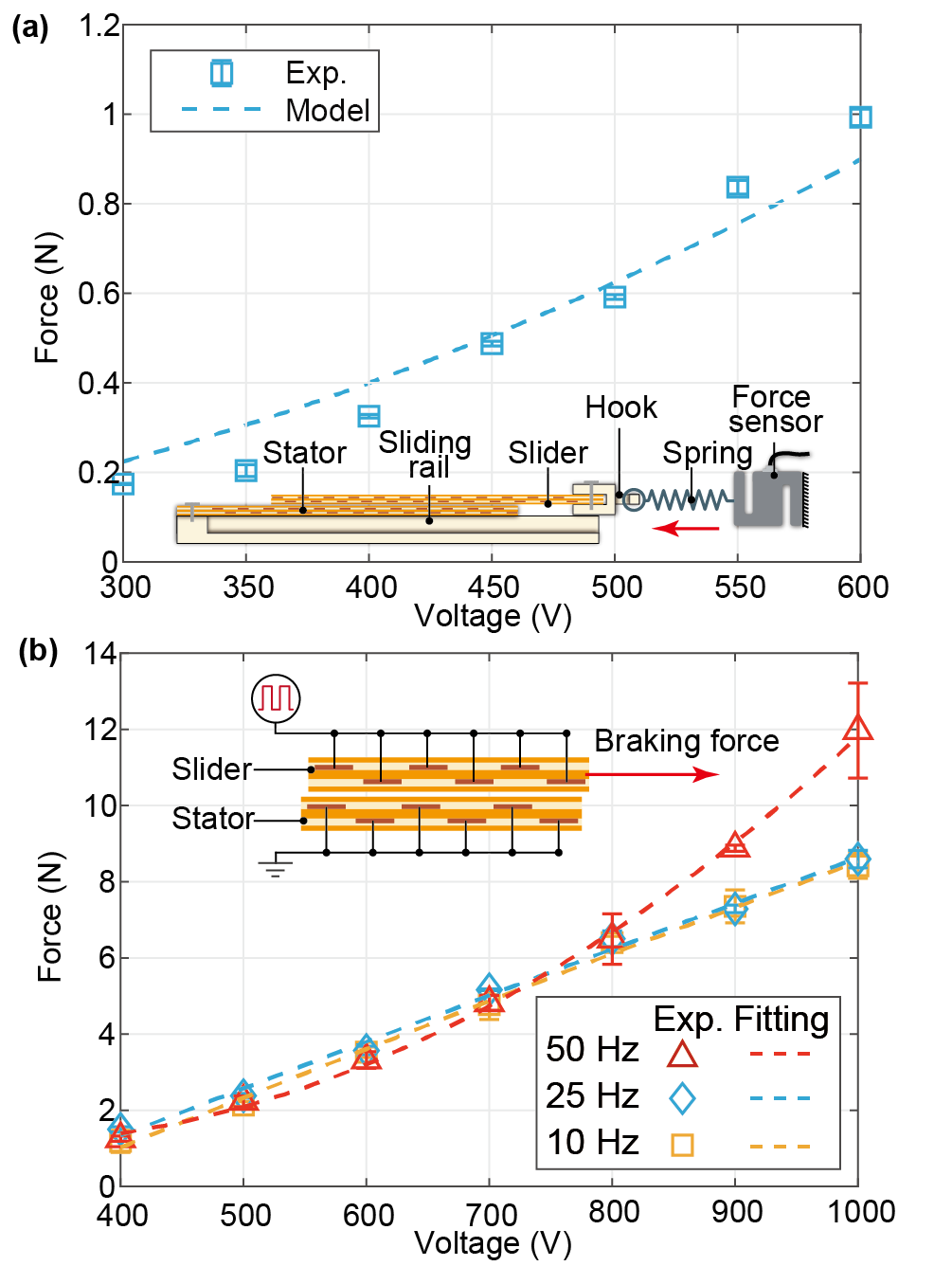}
  \setlength{\abovecaptionskip}{-0.1cm}
  \caption{The actuation force and braking force of the actuator. (a) Schematic of the experimental setup and results of the actuation force for different voltage amplitudes. (b) Results of the braking force for different voltage amplitudes and driving frequencies.}
  \label{fig8}
\end{figure}

In the actuation force measurement setup shown in Fig.~\ref{fig8}a, the stator was fixed on a 3D-printed sliding rail through screws. The slider was connected to a force sensor (LSB~201, FUTEK) via a hook and a spring. During the test, the actuator was applied with sinusoidal voltages with a 10~Hz driving frequency. As the slider stretched the spring leftward, the force sensor recorded the increasing force in real-time. The maximum force value was recorded as the actuation force since the slider could no longer extend the spring at a given voltage amplitude. The relationship between actuation force and voltage amplitude is shown in Fig.~\ref{fig8}a, which is in agreement with the finite element model. A maximum actuation force of 1.02~N (corresponding to an actuation stress of approximately 241~N/m\textsuperscript{2}) was achieved at 600~V$_\text{0-p}$ voltage amplitude.

\begin{figure}[htbp]
  \centering
  \includegraphics[width=8cm]{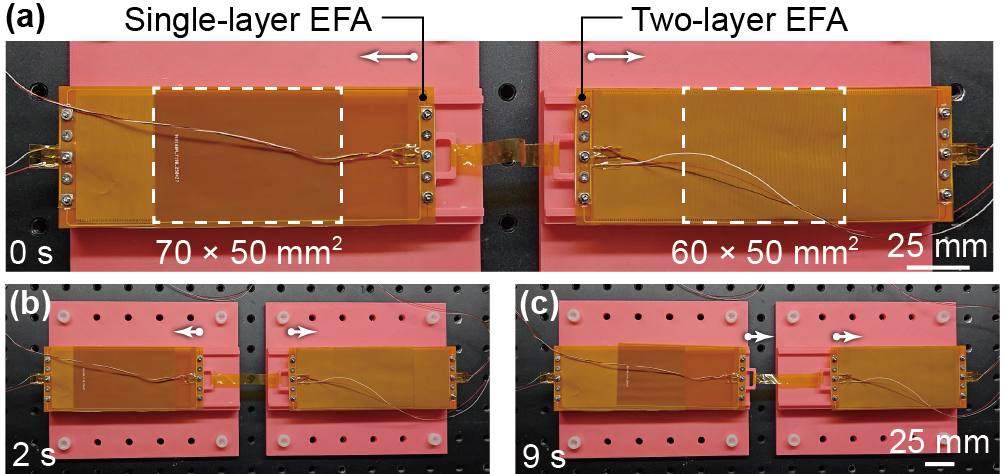}
  \setlength{\abovecaptionskip}{-0.1cm}
  \caption{The demonstrations of a tug-of-war between a single-layer EFA and the proposed two-layer EFA. (a) Initial state: the single-layer actuator with a larger initial effective area. (b) The tape is tighten as the two actuators start to compete. (c) The two-layer actuator eventually pulls the single-layer actuator away.}
  \label{fig9}
  \vspace{-0.5cm}
\end{figure}
\subsection{Results of braking force}
The experimental setup for evaluating the electrostatic adhesion braking force is shown in Fig.~\ref{fig8}b. The slider was connected to a force sensor (LSB203, MTS) and moved at a constant velocity of 50~mm/min by a tensile test platform (Model~42, MTS). During testing, all electrodes on the slider were connected to a bipolar square-wave AC voltage, while the stator electrodes were grounded. The relationships between braking force and voltage amplitude under different driving frequencies are shown in Fig.~\ref{fig8}b. The results indicate that the braking forces at 10~Hz and 25~Hz were similar; however, at 50~Hz, the braking force increased significantly at high voltage amplitude, reaching a maximum value of approximately 12.0~N at $\pm1000$~V. It is noteworthy that repeated vertical tensile tests may cause glass beads to detach from between the two films, increasing the friction coefficient. Therefore, the braking force measured in the tensile tests may be higher than that achievable under normal operating conditions.


\section{DEMONSTRATIONS}
\subsection{Tug-of-war}
To visually compare the actuation forces between conventional single-layer EFA and the two-layer EFA, a tug-of-war platform was designed as shown in Fig.~\ref{fig9} and Movie. On the left side, a single-layer actuator with an initial effective area of $70 \times 50~\text{mm}^2$ was placed, while a two-layer actuator with an initial effective area of $60 \times 50~\text{mm}^2$ was positioned on the right side. The sliders of both actuators were connected via a polyimide tape. Both actuators were simultaneously driven by sinusoidal voltages at 500~V$_\text{0-p}$ amplitude and 5~Hz driving frequency.

At the beginning of the competition, the tape remained slack. After approximately 2 seconds, the tape was pulled taut, and both actuators began exerting opposing forces. Then, the single-layer actuator was pulled off-axis and lost the competition. By 9 seconds, the two-layer actuator had dragged the single-layer actuator out.

\begin{figure}[htbp]
  \centering
  \includegraphics[width=8cm]{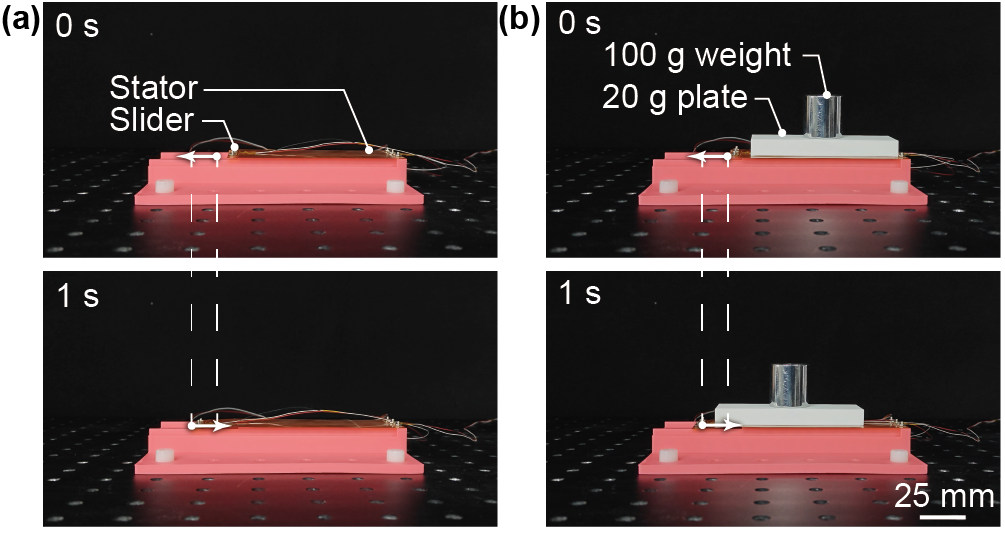}
  \setlength{\abovecaptionskip}{-0.1cm}
  \caption{The demonstrations of the payload capacity of the actuator. The slider of the actuator operates (a) no-load and (b) with a payload of approximately 68.18 times its own weight.}
  \label{fig10}
  \vspace{-0.5cm}
\end{figure}

\subsection{Payload operation}
To evaluate the payload capability of the two-layer EFA, a testing platform was constructed as shown in Fig.~\ref{fig10}. The stator was mounted on a sliding rail, while the slider was placed on top of the stator. When driven by sinusoidal voltages at 500~V$_\text{0-p}$ amplitude and 30~Hz driving frequency, the slider performed a reciprocating motion, as shown in Fig.~\ref{fig10}a and Movie for the unloaded case.

Subsequently, a 3D-printed plate (20~g) and an additional mass (100~g) were placed atop the slider, resulting in a total payload equivalent to approximately 68.2 times the actuator’s own weight. As shown in Fig.~\ref{fig10}b, the actuator maintained its full stroke without observable degradation in motion performance under this loading condition.

\subsection{One-DOF robotic arm}
To demonstrate the integrated braking capability of the two-layer EFA, the actuator was implemented as both the actuation and braking mechanism for a one-DOF robotic arm. Benefiting from the force superposition capability of the two-layer design, the actuator was constructed using three stator films and two slider films. The stator was mounted on the upper arm, while the slider was connected to the forearm via a thin cable. The actuator was driven by voltages at 500~V$_\text{0-p}$ voltage amplitude and 5~Hz driving frequency.

\begin{figure}[htbp]
  \centering
  \includegraphics[width=8cm]{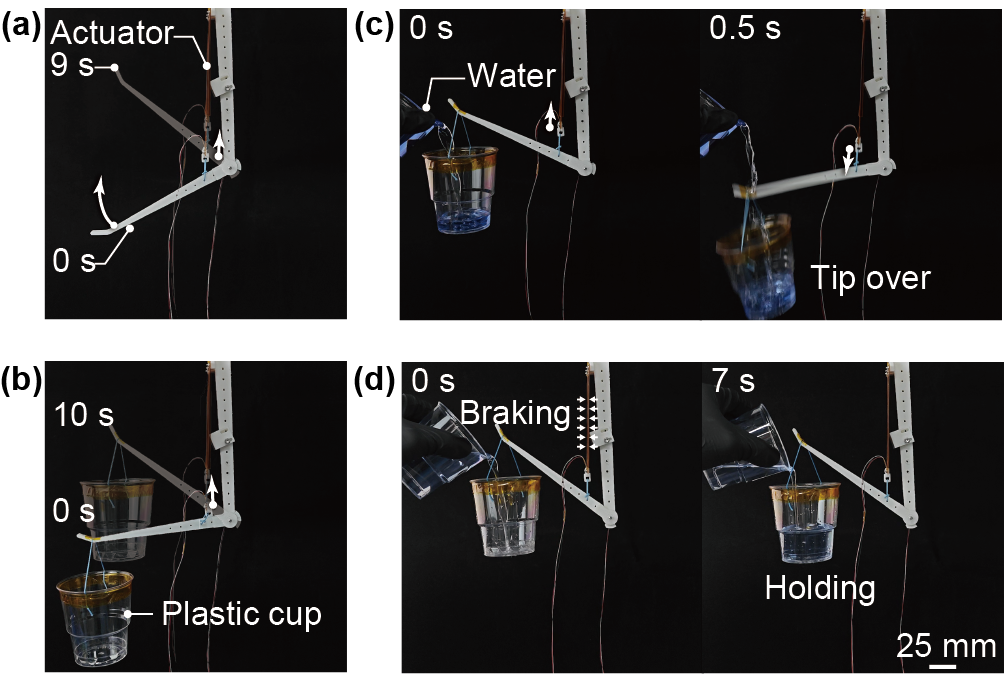}
  \setlength{\abovecaptionskip}{-0.1cm}
  \caption{The demonstrations of the actuator for a one-DOF robotic arm. The robotic arm lifts (a) no-load or (b) with a plastic cup. (c) The actuator slips backward when the water exceeds 20 g, causing the cup to tip over. (d) The actuator switches to the braking mode at the target position and is able to hold water exceeding 60 g.}
  \label{fig11}
\end{figure}

As shown in Fig.~\ref{fig11}a and Movie, the actuator successfully drove the robot arm in elbow-like rotational motion under unloaded conditions. The arm was also able to lift a plastic cup with a mass of 11.2~g from a lower initial angle, as illustrated in Fig.~\ref{fig11}b. To further evaluate the payload limit during dynamic motion, water was incrementally added into the cup during operation. As shown in Fig.~\ref{fig11}c, when the added water mass exceeded approximately 20~g at 1.5~seconds, the actuation force was no longer sufficient to maintain motion, resulting in slipping backward and ultimately causing the cup to tip over at 0.5~seconds.

After reaching the target position by actuation mode, the actuator was switched to braking mode by applying a $\pm1000$~V square-wave voltage at 50~Hz to the slider electrodes while grounding the stator electrodes. Under braking mode, as shown in Fig.~\ref{fig11}d, the actuator was able to hold the arm position even when the payload exceeded 60~g at 7~seconds, preventing tipping and maintaining stability.

\begin{figure}[htbp]
  \centering
  \includegraphics[width=8cm]{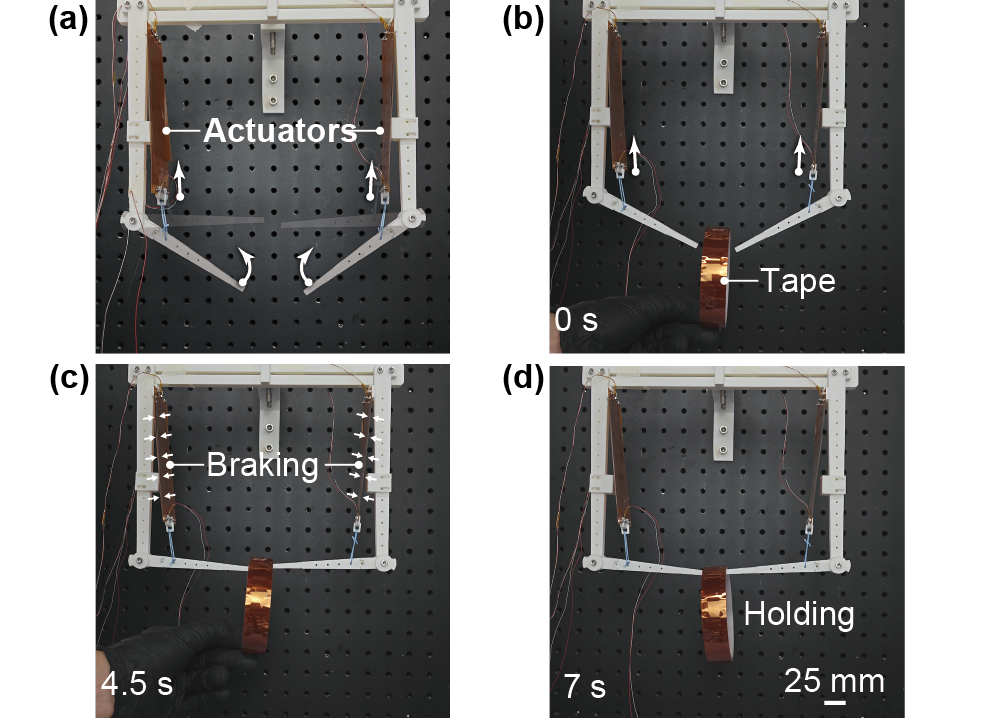}
  \setlength{\abovecaptionskip}{-0.1cm}
  \caption{A dual-mode gripper performs object grasping through the actuation mode of the actuator, and holds the object after closure using the braking mode.}
  \label{fig12}
  \vspace{-0.5cm}
\end{figure}

\subsection{Dual-mode gripper}
Based on two one-DOF robot arms, a dual-mode gripper was constructed by modifying the end-effector structures and assembling them into a gripping mechanism, as illustrated in Fig.~\ref{fig12} and Movie. The gripper successfully grasped a polyimide tape weighing approximately 20~g. During the grasping process, both actuators operated in actuation mode (500~V$_\text{0-p}$ voltage amplitude with 5~Hz driving frequency) to drive the gripper fingers inward, allowing insertion into the hollow center of the tape. After approximately 4.5~seconds, once the gripper fully closed, the actuators were switched to braking mode (applying $\pm1000$~V square-wave voltage at 50~Hz to the slider electrodes while grounding the stator electrodes), thereby holding the grasped object.


\section{CONCLUSIONS}

In summary, a two-layer EFA with integrated braking capability was proposed and experimentally demonstrated to address the force limitations of conventional EFAs operating in air. The two-layer electrode design enables a smaller effective electrode pitch without requiring higher fabrication precision, thereby significantly enhancing the actuation stress. Finite element simulations were conducted to analyze the effects of design parameters and fabrication misalignment on actuator performance. Experimental results confirmed that the proposed actuator achieved high actuation stress, precise positioning, and substantial payload capacity. A series of demonstrations, including tug-of-war, payload operation, robotic arm manipulation, and dual-mode gripping, further validated its applicability in both actuation and braking modes. Currently, the integrated braking mechanism is affected by the glass beads during operation, which may significantly reduce the braking force. Future improvements may involve encapsulation of the actuator in a dielectric liquid to ensure stable braking performance. Additionally, the present fabrication process employs a relatively thick base layer, which could be further optimized to enable the two-layer design to achieve even higher actuation stresses. This two-layer EFA provides a promising solution for lightweight robotic systems that require compact, precise actuation combined with integrated braking functionality.

\balance
\bibliographystyle{IEEEtran}

\bibliography{ref}  
\end{document}